\documentclass[letterpaper, 10pt, conference]{ieeeconf}
\usepackage{times}
\usepackage[numbers]{natbib}
\usepackage{url}
\usepackage{graphicx}
\usepackage[table]{xcolor}
\usepackage{caption}
\usepackage{subcaption}
\usepackage{amssymb}
\usepackage{amsmath}
\usepackage{booktabs}
\usepackage{stfloats}

\IEEEoverridecommandlockouts
\usepackage{tabularx}

\title{\Large \bf
Decoupling Vision, Language, and Action for Efficient Multi-Task Robot Policies
}

\author{Xiatao Sun$^{1*}$, Chen Liang$^{1}$, Ziyao Zeng$^{1}$, Qian Wang$^{1}$, \\
Haoyang Zhang$^{2,3}$, Yue Sun$^{3}$, Qiucheng Li$^{3}$, and Daniel Rakita$^{1}$%
\thanks{$^{1}$Department of Computer Science, Yale University, New Haven, CT, USA.}%
\thanks{$^{2}$Peking University, Beijing, China.}%
\thanks{$^{3}$Digients, Singapore.}%
\thanks{$^{*}$Corresponding author: {\tt\small xiatao.sun@yale.edu}}%
}

\begin{document}

\maketitle
\thispagestyle{empty}
\pagestyle{empty}
\bstctlcite{IEEEexample:BSTcontrol}

\begin{abstract}
Vision-Language-Action (VLA) policies commonly run Vision-Language
Model (VLM) backbones with billions of parameters at every policy
inference, which costs latency and energy. We revisit a decoupled
alternative for multi-task manipulation: separate vision and language
encoders whose representations condition a compact action head. We run
a standardized comparison that varies the vision encoder, the language
encoder, and the action head while holding the demonstrations, the training-step budget, the tasks, the evaluation protocol, and the
measurement platform fixed, against seven VLA baselines. The resulting
Decoupled Embodiment Model (DEM) combines a fine-tuned DINOv3 vision
encoder, a frozen NeoBERT language encoder, and a MeanFlow head that
generates an action chunk in one forward pass. On 18 RoboCasa tasks
evaluated with held-out instruction paraphrases and randomized scenes,
DEM reaches 55.6\% mean success against 56.9\% for GR00T N1.7 and
54.6\% for $\pi_{0.5}$, and on three real-robot tasks it reaches
66.0\% against 68.0\% for GR00T N1.7. On the same workstation, DEM needs 6.1\,ms per policy forward pass, a maximum throughput of 162.7 policy calls per second, and draws an estimated 2.07\,J of GPU energy per call, eight to seventeen times the throughput and six to fifteen times less energy than these VLM-backbone policies. Within this
trained-task regime, DEM sits on the observed success--latency--energy
frontier and provides a strong, efficient baseline for
language-conditioned robot skills.
\end{abstract}

\section{Introduction}
\label{sec:introduction}

Most Vision-Language-Action (VLA) models build on a large
Vision-Language Model (VLM) and attach an action
module \cite{kim2024openvla,black2024pi0,bjorck2025gr00tn1}. Running a
backbone with billions of parameters at every policy inference costs
latency and energy: $\pi_{0.5}$ \cite{black2025pi05}, for example,
needs on the order of $100$\,ms per inference. Action
chunking \cite{zhao2023act} and asynchronous
execution \cite{black2025rtc} accommodate slow inference during
execution, but they do not shorten the perception-to-action computation
itself. These costs matter most for low-level manipulation, where
policies must stay reactive under energy constraints and
high-frequency control runs at 25--50\,Hz \cite{kim2025oft}.

For such policies, the VLM mainly provides representations of the
images and the instruction that condition action prediction. That role
does not require generating language, which is what a decoder-only VLM
is built and sized for. Autoregressive VLAs such as OpenVLA run the
decoder only to emit action tokens, and most recent VLAs delegate
action generation to a separate module. The split between encoder-only
language models for understanding \cite{devlin2019bert} and
decoder-only models for generation \cite{brown2020gpt3} suggests an
alternative: separate vision and language encoders that supply the
representations an action head conditions on.

Earlier generalist policies used this decoupled design.
RT-1 \cite{brohan2022rt1} paired an EfficientNet with a frozen sentence
encoder in a 35M-parameter policy, and Octo \cite{octo2024octo}
conditioned a small transformer on frozen T5 embeddings. Both relied on the pretrained components of their time, and as pretraining investment moved toward decoder-only models, robotics adopted the strongest
available VLM checkpoints \cite{black2024pi0}. Standalone encoders have
since advanced, with DINOv3 \cite{simeoni2025dinov3} and
NeoBERT \cite{lebreton2025neobert} among them, and, as
Section~\ref{sec:detour} reviews, they now produce representations
competitive with those of models several times their size.

\begin{figure}[t]
  \centering
  \includegraphics[width=\columnwidth]{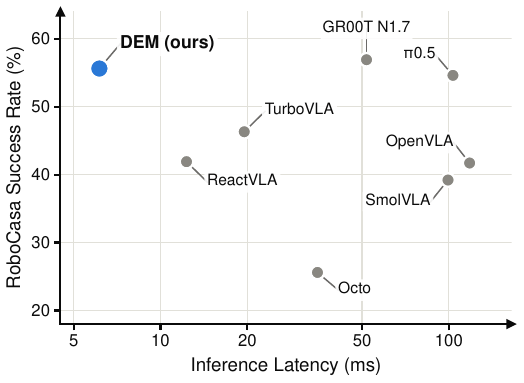}
    \caption{\textbf{Success against forward-pass latency.} RoboCasa success rate (18 tasks, 100 rollouts each) against median forward-pass latency on one RTX PRO 6000, log scale, for DEM and the seven baselines. DEM reaches the success range of the VLM-backbone policies GR00T N1.7 and $\pi_{0.5}$ at 6.1\,ms per call, eight to seventeen times less latency.}
  \label{fig:teaser}
\end{figure}

Whether better encoders translate into better policies is an empirical question. A jointly pretrained VLM may provide representations that separate encoders lack, and its capacity also carries a substantial inference cost. We therefore ask a deliberately scoped question: when a policy is trained to execute a fixed collection of language-conditioned manipulation skills, how much observed success does replacing a jointly fused VLM backbone with modern standalone encoders retain, and what inference cost does it save? We answer the question with a study of the vision encoder, the language encoder, and the action head that holds the demonstrations, the training steps, the tasks, the evaluation protocol, and the measurement platform fixed, and with a system-level comparison against seven VLM-backbone and decoupled baselines.

The study yields the Decoupled Embodiment Model (DEM), which pairs a
fine-tuned DINOv3 vision encoder and a frozen NeoBERT language encoder
with a MeanFlow action head \cite{geng2025meanflow}. MeanFlow generates
a whole action chunk in a single forward pass and accounts for a large
share of DEM's inference speed (Section~\ref{sec:results_action}). On
18 RoboCasa tasks with held-out language paraphrases and randomized
scenes, DEM reaches 55.6\% success, against 56.9\% and 54.6\% for VLM-backbone policies seven to nine times its size, while running inference eight to seventeen times faster and drawing six to fifteen times less energy per inference. We also evaluate it on
three real-robot tasks.

This paper makes three contributions.
\begin{itemize}
  \item A standardized system-level evaluation of seven VLM-backbone
    and decoupled robot policies on the same 18-task demonstration set,
    training-step budget, evaluation protocol, and measurement
    platform, scoped to trained skills with held-out language
    paraphrases and randomized scenes.
  \item A component study of decoupled policies showing that, under
    this protocol, fine-tuning the vision encoder has a far larger
    observed effect than fine-tuning the language encoder, and that the
    resulting representation stays effective across regression-,
    diffusion-, and flow-based action heads.
  \item DEM, an efficient configuration on the observed success--latency--energy frontier, with a measured 6.1\,ms forward pass and an estimated 2.07\,J of GPU energy per policy call, released as open code\footnote{\url{https://github.com/Apollo-Lab-Yale/decoupled-embodiment-model}}.
\end{itemize}

\section{Related Work}
\label{sec:detour}

\textbf{Decoupled policies and VLM backbones.}
\label{sec:detour_decoupled}
The first language-conditioned generalist policies kept separate
encoders: RT-1 \cite{brohan2022rt1} paired an EfficientNet with a
frozen sentence encoder, Octo \cite{octo2024octo} conditioned a small
transformer on frozen T5 embeddings, and BAKU \cite{haldar2024baku} put
interchangeable action heads behind separate encoders. Their components
dated from 2019: encoder-only language modeling saw little follow-up
after RoBERTa \cite{warner2024modernbert}, and no pretrained visual
representation of the time worked across embodied
tasks \cite{majumdar2023vc1}. As pretraining investment moved to
decoder-only generation \cite{brown2020gpt3}, robotics repurposed VLMs
as embedding providers, and the recipe settled into
OpenVLA \cite{kim2024openvla}, the $\pi$
series \cite{black2024pi0,black2025pi05}, and
GR00T \cite{bjorck2025gr00tn1}. The backbone's latency is handled by
action chunking \cite{zhao2023act}, parallel decoding \cite{kim2025oft},
real-time chunk stitching \cite{black2025rtc}, smaller or truncated
backbones \cite{wen2024tinyvla,shukor2025smolvla}, and token caching or
early exit \cite{xu2025vlacache,yue2024deervla}, all of which keep the decoder-only structure. A separate line makes compact policies cheaper without a VLM, through low-rank or geometry-structured training of diffusion policies \cite{sun2025dynamic,sun2026hybrid}, pose- or mesh-based observations \cite{sun2025prismdp,wang2025subsecond}, learned viewpoint selection \cite{sun2024optimizing}, and attention shaping against shortcut learning \cite{sun2026artificial}.

\textbf{Encoder revival.}
\label{sec:detour_revival}
The components that limited the decoupled policies have since improved.
DINOv3 \cite{simeoni2025dinov3} features exceed the language-supervised
encoders that current VLMs use as vision towers on dense tasks and match
them on classification. ModernBERT \cite{warner2024modernbert},
NeoBERT \cite{lebreton2025neobert}, and mmBERT \cite{marone2025mmbert}
renewed encoder pretraining, and paired-training experiments that hold
data, architecture, and parameter count fixed show encoders beating
decoders on representation tasks, with a 150M encoder above a 400M
decoder on MNLI \cite{weller2026ettin}. Inside VLMs, with the language
model held fixed, perceptual performance is set by the vision
tower \cite{tong2024cambrian}. Whether a policy built on these encoders reaches the success of VLM-backbone policies is the question this paper tests.

\textbf{Concurrent work.}
TurboVLA \cite{xie2026turbovla} and ReactVLA \cite{guo2026reactvla}
arrive at a similar design and evaluate it on LIBERO, where a
0.54M-parameter policy conditioned on a task index reaches 95.1\% \cite{sendai2026minerva}, so parity there separates methods poorly. MeanFlow \cite{geng2025meanflow} has entered manipulation through per-task point-cloud policies \cite{sheng2025mp1} and, concurrently with our work, as the action expert of language-conditioned policies \cite{chen2026meanflowvla,guo2026reactvla}; one-step action generation for VLAs \cite{chen2026letitbesimple} and asynchronous per-modality processing for reactive control \cite{vanjani2026damvla} are also concurrent. Our contribution is complementary: a standardized comparison on RoboCasa \cite{nasiriany2024robocasa} and a physical system that varies the vision encoder, the language encoder, and the action head under one protocol, with TurboVLA and ReactVLA as baselines.

\section{DEM: Decoupled Embodiment Model}
\label{sec:method}

\begin{figure*}[t]
  \centering
  \includegraphics[width=\textwidth]{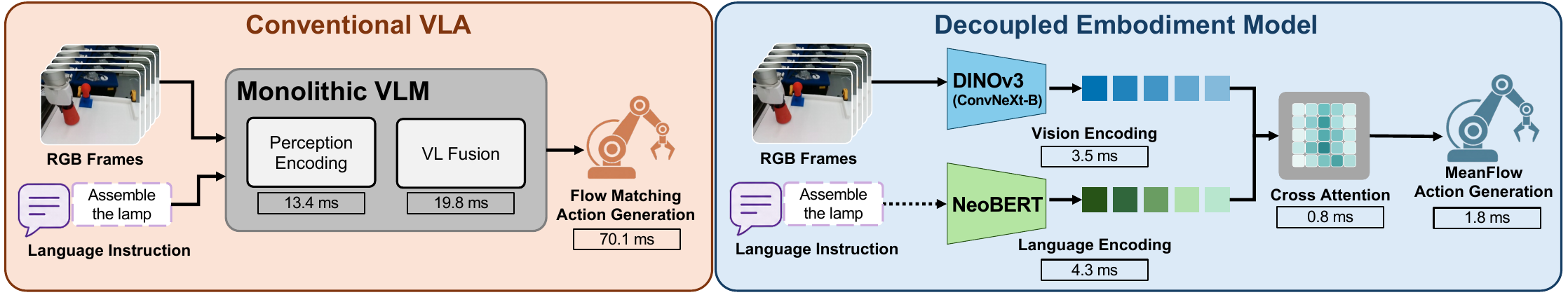}
  \caption{\textbf{Conventional VLA versus DEM.} Left: a conventional VLA ($\pi_{0.5}$ \cite{black2025pi05}) routes images and instruction through one VLM and generates actions by multi-step flow matching; in this implementation the instruction is re-encoded at every step. Right: DEM encodes images with DINOv3 ConvNeXt-B and the instruction with NeoBERT, fuses them by cross-attention inside the action head, and generates the chunk with MeanFlow in one forward pass. The dashed language pathway runs only when the instruction changes. Boxes give per-stage latencies and sum to the figures of Table~\ref{tab:efficiency}.}
  \label{fig:architecture}
\end{figure*}

DEM does not share a backbone across modalities. A vision encoder, a
language encoder, and an action head are three separate networks whose
outputs meet only in the cross-attention inside the head
(Fig.~\ref{fig:architecture}). The specific modules, DINOv3 ConvNeXt-B for vision, NeoBERT for language, and
MeanFlow for the action head, were selected by the controlled study of
Sections~\ref{sec:results_vision} to \ref{sec:results_action}.

\subsection{Vision encoder}
\label{sec:method_vision}

Each camera frame passes through a DINOv3
ConvNeXt-B \cite{simeoni2025dinov3}. We take the stage-3 feature map, a
$16 \times 16$ grid of 512-dimensional features for a 256-pixel input,
and flatten it into $256$ tokens per camera; with a scene camera and a
wrist camera the vision pathway emits $N_v = 512$ tokens per control
step. The encoder is initialized from the public DINOv3 checkpoint and
fine-tuned together with the head, which the ablation in
Section~\ref{sec:results_vision} supports.

\subsection{Language encoder}
\label{sec:method_language}

The instruction goes through NeoBERT \cite{lebreton2025neobert}, an encoder-only language model with 222M parameters as loaded. We pad or truncate
the instruction to $N_l = 32$ tokens and keep the final-layer hidden
state of every token with its attention mask rather than a pooled sentence vector. NeoBERT stays frozen, since fine-tuning it did not change success in our experiments (Section~\ref{sec:results_language}). Because the language tokens are computed without reference to the images, the language pathway is optional at each step: when the instruction has not changed, the policy reuses the cached tokens, and when a new command arrives, it re-encodes. In the jointly fused VLM implementations evaluated here, language and image tokens pass through the shared backbone together at every step, so the language-pathway computation cannot be reused the way DEM reuses its independently encoded language tokens.

\subsection{Action head}
\label{sec:method_head}

The head generates a chunk of $H$ actions
$\mathbf{a} \in \mathbb{R}^{H \times A}$, with $A$ the action dimension,
as a MeanFlow model \cite{geng2025meanflow}. Flow matching learns the instantaneous velocity $v(\mathbf{z}_t, t)$ of the
interpolation
$\mathbf{z}_t = (1-t)\,\mathbf{a} + t\,\boldsymbol{\epsilon}$,
$\boldsymbol{\epsilon} \sim \mathcal{N}(\mathbf{0}, \mathbf{I})$, and
recovers $\mathbf{a}$ from noise by integrating $v$ over many small
steps. MeanFlow instead learns the average velocity over an interval
$[r, t]$,
\begin{equation}
  \mathbf{u}(\mathbf{z}_t, r, t) = \frac{1}{t - r}\int_r^t v(\mathbf{z}_\tau, \tau)\,\mathrm{d}\tau ,
  \label{eq:avg_vel}
\end{equation}
so that the displacement across the interval is a single product,
$\mathbf{z}_r = \mathbf{z}_t - (t - r)\,\mathbf{u}(\mathbf{z}_t, r, t)$.
Differentiating Eq.~\eqref{eq:avg_vel} with respect to $t$ gives the
MeanFlow identity
\begin{equation}
  \mathbf{u}(\mathbf{z}_t, r, t) = v(\mathbf{z}_t, t) - (t - r)\,\frac{\mathrm{d}}{\mathrm{d}t}\,\mathbf{u}(\mathbf{z}_t, r, t),
  \label{eq:identity}
\end{equation}
which the network $\mathbf{u}_\theta$ is trained to satisfy. Along the
interpolation $v = \boldsymbol{\epsilon} - \mathbf{a}$ is known in
closed form, and the total derivative
$\frac{\mathrm{d}}{\mathrm{d}t}\mathbf{u}_\theta = v\,\partial_{\mathbf{z}}\mathbf{u}_\theta + \partial_t\mathbf{u}_\theta$
costs one Jacobian-vector product. The training loss is
\begin{equation}
  \mathcal{L} = \mathbb{E}\Big[\, w\,\big\|\, \mathbf{u}_\theta(\mathbf{z}_t, r, t \mid \mathbf{C}) - \mathrm{sg}\big(v - (t - r)\tfrac{\mathrm{d}}{\mathrm{d}t}\mathbf{u}_\theta\big) \big\|_2^2 \Big],
  \label{eq:loss}
\end{equation}
where $\mathrm{sg}$ stops gradients through the target, $\mathbf{C}$
is the conditioning context of Sec.~\ref{sec:method_fusion}, and
$w = (\|\cdot\|_2^2 + c)^{-p}$ is the adaptive weight
of \cite{geng2025meanflow} with $c = 10^{-3}$ and $p = 0.5$, which that paper reports as competitive with its default $p = 1$. The pair $(r, t)$ is drawn from a logit-normal distribution and set equal with probability $0.5$, against $0.75$ in the MeanFlow default, in which case Eq.~\eqref{eq:loss} reduces to the flow-matching loss and anchors training.

At inference the head runs once. Starting from
$\boldsymbol{\epsilon} \sim \mathcal{N}(\mathbf{0}, \mathbf{I})$,
\begin{equation}
  \hat{\mathbf{a}} = \boldsymbol{\epsilon} - \mathbf{u}_\theta(\boldsymbol{\epsilon}, 0, 1 \mid \mathbf{C}),
  \label{eq:onestep}
\end{equation}
which is the displacement identity with $r = 0$ and $t = 1$.

\subsection{Assembling the modules}
\label{sec:method_fusion}

The three modules meet inside the head as one set of context tokens.
Let $\mathbf{V} \in \mathbb{R}^{N_v \times d_v}$ be the vision tokens,
$\mathbf{L} \in \mathbb{R}^{N_l \times d_l}$ the language tokens with
mask $\mathbf{m} \in \{0, 1\}^{N_l}$, and
$\mathbf{q} \in \mathbb{R}^{d_p}$ the proprioceptive state. Each is
projected to the head width $d$ and tagged with a learned modality
embedding,
\begin{equation}
  \mathbf{C} = \big[\, \mathbf{V}\mathbf{W}_v + \mathbf{e}_v \;\big\|\; \mathbf{L}\mathbf{W}_l + \mathbf{e}_l \;\big\|\; \mathbf{q}^{\top}\mathbf{W}_p + \mathbf{e}_p \,\big],
  \label{eq:context}
\end{equation}
where $\|$ concatenates along the token axis, so $\mathbf{C} \in \mathbb{R}^{(N_v + N_l + 1) \times d}$ holds $545$ context tokens per step in our configuration. The noisy chunk enters as
$H$ action tokens,
$\mathbf{h}^{(0)} = \mathbf{z}_t\mathbf{W}_a + \mathbf{P}$ with a
learned position embedding $\mathbf{P} \in \mathbb{R}^{H \times d}$,
and the interval $(r, t)$ enters through a time embedding
$\mathbf{c} = \phi_r(r) + \phi_t(t)$. Each of the $D$ blocks applies
self-attention among the action tokens and a feed-forward layer, both
modulated by adaptive layer normalization from $\mathbf{c}$ as in
DiT \cite{peebles2023dit}, and then reads the context:
\begin{align}
  \tilde{\mathbf{h}} &= \mathrm{AdaLNBlock}\big(\mathbf{h}^{(i)}, \mathbf{c}\big), \label{eq:adaln} \\
  \mathbf{h}^{(i+1)} &= \tilde{\mathbf{h}} + \mathbf{g}_i \odot \mathrm{CrossAttn}\big(\mathrm{LN}(\tilde{\mathbf{h}}),\, \mathbf{C},\, \mathbf{m}\big), \label{eq:cross}
\end{align}
where the action tokens are the queries, the context tokens are the keys and values, $\mathbf{m}$ is extended with ones over the vision and proprioceptive tokens so that only padded language tokens are masked, and $\mathbf{g}_i \in \mathbb{R}^{d}$ is a per-channel gate initialized at zero, so each block starts as an unconditional
denoiser and learns how much of the scene and the instruction to read.
A final layer normalization and a zero-initialized linear layer map
$\mathbf{h}^{(D)}$ to $\mathbf{u}_\theta \in \mathbb{R}^{H \times A}$.

In our configuration $d = 768$, $D = 8$, and the head has 107M
parameters. The head is the only module that consumes $\mathbf{C}$, so
swapping either encoder changes nothing but $\mathbf{W}_v$ or
$\mathbf{W}_l$, and every variant in
Sections~\ref{sec:results_vision} to \ref{sec:results_action} plugs into the same head. The vision encoder is fine-tuned from its public checkpoint, the language encoder stays frozen, and the head trains from scratch.

\section{Experimental Setup}
\label{sec:setup}

\begin{figure*}[t]
  \centering
  \includegraphics[width=0.96\textwidth]{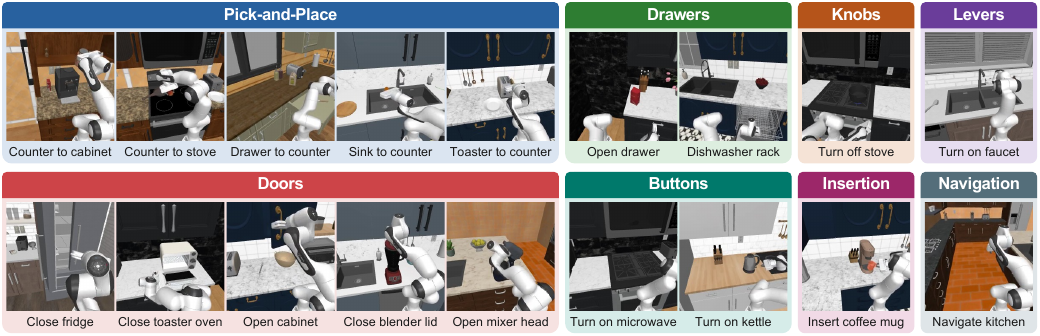}
  \caption{\textbf{Simulated tasks.} One demonstration frame per task, grouped by the skill categories of Table~\ref{tab:main}.}
  \label{fig:tasks}
\end{figure*}

\textbf{Simulation setup.} All simulated experiments run on 18 atomic
tasks in the RoboCasa kitchens \cite{nasiriany2024robocasa}
(Fig.~\ref{fig:tasks}), the tasks for which the NVIDIA
PhysicalAI-Robotics-Manipulation-Kitchen-Demos
dataset \cite{nvidia2025kitchendemos} provides demonstrations; they span
the eight skill categories of the RoboCasa paper and add tasks on
appliances (toaster oven, blender, stand mixer, dishwasher, kettle)
beyond the original atomic-task list. Training uses 500 episodes per task. The dataset ships one instruction per task; to avoid training on a single string, we generate 50 paraphrases of each instruction with Claude Opus 4.8 \cite{anthropic2026claude} and sample one per training episode for every policy. At evaluation, every rollout uses one of 10 further paraphrases that never appear in training. Evaluation layouts and object placements are drawn from RoboCasa's randomization and are not those of the demonstrations. Every policy and every variant is trained for 300k steps on the same demonstrations and evaluated over 100 rollouts per task. With 100 rollouts, the binomial standard error of a task's success rate is at most 5 points and that of the 18-task mean about 1.2 points. This figure covers evaluation sampling only, since each configuration is trained once; we report success differences descriptively.

\textbf{Baselines.} We compare against OpenVLA \cite{kim2024openvla},
Octo \cite{octo2024octo}, SmolVLA \cite{shukor2025smolvla},
$\pi_{0.5}$ \cite{black2025pi05}, GR00T N1.7 \cite{nvidia2026gr00tn17},
TurboVLA \cite{xie2026turbovla}, and ReactVLA \cite{guo2026reactvla}, all fine-tuned on the same demonstrations with their official implementations and default recipes, updating all parameters. DEM and the chunked baselines generate 16 actions per inference and replan every eight control steps. Each baseline keeps its native fusion, action module, and training recipe, so the comparison is system-level.

\textbf{Compute.} Training runs on eight H100 GPUs. Evaluation and
every latency and power measurement run on one workstation with an RTX
PRO 6000 GPU, an AMD Ryzen Threadripper PRO 9965WX CPU, and 128\,GB of memory. Latency covers the policy's forward pass at batch size 1, reported as the median over the timed calls after 100 warm-up calls with device synchronization before each timestamp, and excludes camera capture, preprocessing, and robot communication, which are common to all policies; energy is the mean GPU power during inference, sampled with NVML at 10\,Hz, times the median latency, with idle draw not subtracted. It is a device-level estimate of GPU energy per policy call, excludes the CPU, cameras, communication, and actuation, and serves for comparisons on this shared platform. Inference runs in BF16 on the GPU at its factory power limit with unchanged clock settings.

\textbf{Real-world setup.} Real-world experiments use an xArm~7 mounted
on a linear rail driven by a linear motor, which together form an
eight-degree-of-freedom manipulation system. We evaluate three
FurnitureBench \cite{heo2023furniturebench} assembly tasks, drawer,
lamp, and cabinet, with 200 teleoperated episodes per task, 100k training steps for every policy, the same paraphrase protocol, and 50 rollouts per task and method (standard error up to 7 points per task, about 4 points for the three-task mean).
On the robot we compare against GR00T N1.7, the strongest VLM-backbone
policy in simulation, $\pi_{0.5}$, the most widely deployed one and the
architecture of Fig.~\ref{fig:architecture}, and TurboVLA, the
strongest concurrent decoupled policy; the remaining baselines score
below TurboVLA in simulation while spending more energy per inference
(Tables~\ref{tab:main} and \ref{tab:efficiency}).

\section{Results}
\label{sec:results}

\subsection{Overall benchmark}
\label{sec:results_main}

Table~\ref{tab:main} reports success rates on the 18 RoboCasa atomic
tasks. DEM averages 55.6\%, against 56.9\% for GR00T N1.7 and 54.6\% for $\pi_{0.5}$, point estimates that differ by 1.3 and 1.0 points; rollout sampling alone gives the DEM--GR00T difference a 95\% interval of $-1.3 \pm 3.3$ points. Since the experiment does not estimate training-seed variability, we read the three as occupying a similar observed success range rather than as an ordering. What the experiment does resolve is the order-of-magnitude difference in forward-pass cost at this success range (Table~\ref{tab:efficiency}). DEM has 417M parameters, 195M of them active per step, against roughly 3B for either VLM-backbone policy (Table~\ref{tab:efficiency}). It has
the highest success rate on drawers, levers, and buttons, trails GR00T N1.7 on pick-and-place, doors, and insertion,
and has its largest deficit on the knob task (37.0\% against 49.0\%). The two other decoupled policies sit below the VLM-backbone
group, TurboVLA at 46.3\% and ReactVLA at 41.9\%, and Octo, which pairs
2024-era encoders with a diffusion head, is last at 25.6\%.

\begin{table*}[t]
  \centering
  \footnotesize
  \setlength{\tabcolsep}{5pt}
  \caption{\textbf{RoboCasa success rate (\%).} 18 tasks, 100 rollouts each, grouped by the eight RoboCasa skill categories \cite{nasiriany2024robocasa} (task counts in parentheses); a category score is the mean over its tasks and Avg.\ the mean over all 18. The blender lid and stand-mixer head are counted with doors. All policies fine-tuned on the same demonstrations for 300k steps; standard error up to 5 points per task and about 1.2 points for Avg. Bold marks the highest point estimate in each column, not a significant difference.}
  \label{tab:main}
  \begin{tabular}{l cccccccc c}
    \toprule
    Method & Pick-place (5) & Doors (5) & Drawers (2) & Knobs (1) & Levers (1) & Buttons (2) & Insertion (1) & Navigation (1) & Avg.\ (18) \\
    \midrule
    OpenVLA \cite{kim2024openvla} & 24.6 & 47.2 & 51.0 & 34.0 & 45.0 & 61.0 & 51.0 & 38.0 & 41.7 \\
    Octo \cite{octo2024octo} & 13.0 & 32.8 & 30.0 & 25.0 & 35.0 & 28.5 & 20.0 & 35.0 & 25.6 \\
    SmolVLA \cite{shukor2025smolvla} & 27.8 & 43.4 & 44.0 & 25.0 & 48.0 & 54.0 & 44.0 & 37.0 & 39.2 \\
    $\pi_{0.5}$ \cite{black2025pi05} & 34.2 & 55.8 & 72.5 & 44.0 & 60.0 & 78.0 & 63.0 & \textbf{64.0} & 54.6 \\
    GR00T N1.7 \cite{nvidia2026gr00tn17} & \textbf{36.0} & \textbf{60.2} & 75.0 & \textbf{49.0} & 64.0 & 79.0 & \textbf{64.0} & 59.0 & \textbf{56.9} \\
    TurboVLA \cite{xie2026turbovla} & 27.6 & 55.0 & 62.5 & 36.0 & 40.0 & 65.5 & 42.0 & 46.0 & 46.3 \\
    ReactVLA \cite{guo2026reactvla} & 23.8 & 43.4 & 50.5 & 33.0 & 40.0 & 68.0 & 59.0 & 50.0 & 41.9 \\
    \midrule
    DEM (ours) & 33.2 & 56.4 & \textbf{79.5} & 37.0 & \textbf{75.0} & \textbf{81.5} & 63.0 & 56.0 & 55.6 \\
    \bottomrule
  \end{tabular}
\end{table*}

\subsection{Forward-pass latency and estimated GPU energy}
\label{sec:results_efficiency}

Table~\ref{tab:efficiency} reports the per-inference cost of each policy on the workstation of Section~\ref{sec:setup}: median latency, the maximum policy-call throughput it allows (its reciprocal), and the estimated GPU energy of one call. DEM produces an action chunk in 6.1\,ms, a maximum forward-pass throughput of 163 policy calls per second, and spends an estimated 2.07\,J per call. GR00T N1.7 and $\pi_{0.5}$, the two policies in the same success range, reach 19.3 and 9.7 calls per second and spend 13.7 and 31.7\,J. The two other decoupled policies are closer: TurboVLA reaches 51.2 calls per second and 3.06\,J and ReactVLA 81.2 calls per second and 5.82\,J.

The DEM figures use the on-demand language pathway of
Fig.~\ref{fig:architecture}: the instruction is encoded when it arrives
and its tokens are cached, so a control step pays for the vision
encoder, the cross-attention, and one MeanFlow evaluation. The jointly fused VLM implementations evaluated here offer no equivalent reuse (Section~\ref{sec:method_language}). Running
NeoBERT at every step instead raises DEM's latency to 10.4\,ms (96 calls per second) and its energy to 3.09\,J per call, which still leaves it the fastest policy in the table.

\begin{table}[t]
  \centering
  \footnotesize
  \setlength{\tabcolsep}{3pt}
  \caption{\textbf{Per-inference cost} on one RTX PRO 6000 at batch size 1: parameters, median latency, maximum throughput (its reciprocal), and estimated GPU energy per policy call (Section~\ref{sec:setup}). DEM with the instruction cached and with NeoBERT run at every step. Bold marks the lowest cost and highest throughput. $^\dagger$ReactVLA as reproduced by us; its paper reports 390M for the action transformer alone. $^\ddagger$Total; 195M run at every step, NeoBERT (222M) only when the instruction changes.}
  \label{tab:efficiency}
  \begin{tabular*}{\columnwidth}{@{\extracolsep{\fill}} l r r r r @{}}
    \toprule
    Method & Params (M) & Latency (ms) & Calls/s & Energy (J) \\
    \midrule
    OpenVLA & 7541 & 118.0 & 8.5 & 45.13 \\
    Octo & \textbf{202} & 35.0 & 28.6 & 4.88 \\
    SmolVLA & 450 & 99.3 & 10.1 & 17.26 \\
    $\pi_{0.5}$ & 3600 & 103.3 & 9.7 & 31.68 \\
    GR00T N1.7 & 3000 & 51.8 & 19.3 & 13.71 \\
    TurboVLA & 216 & 19.5 & 51.2 & 3.06 \\
    ReactVLA & 2240$^\dagger$ & 12.3 & 81.2 & 5.82 \\
    \midrule
            DEM, cached & 417$^\ddagger$ & \textbf{6.1} & \textbf{162.7} & \textbf{2.07} \\
    DEM, per-step & 417$^\ddagger$ & 10.4 & 96.0 & 3.09 \\
    \bottomrule
  \end{tabular*}
\end{table}

\subsection{Vision encoder ablation}
\label{sec:results_vision}

Table~\ref{tab:abl_vision} swaps the vision encoder with the language
encoder, fusion, and head fixed. Fine-tuning the encoder matters more
than which encoder is chosen: the frozen DINOv3 ConvNeXt-B reaches
32.6\%, the fine-tuned one 55.6\%, and every fine-tuned encoder in the
table lands between 51.8\% and 55.6\%. The two VLM rows use each VLM's vision tower on its own with the same head and language encoder.
Neither reaches a higher success point estimate than DINOv3 under this protocol: they reach 53.9\% and 51.8\% with about
five times the parameters, three times the latency, and four times the
energy per inference.

These rows do not contradict Section~\ref{sec:method_language}. Inside the evaluated VLMs the two towers share one attention stack, so the vision tower cannot run without the language model and the instruction is re-encoded at every step; in DEM each tower writes to a static context (Section~\ref{sec:method_fusion}), so the language tokens can be cached.

\begin{table}[t]
  \centering
  \scriptsize
  \setlength{\tabcolsep}{2pt}
  \caption{\textbf{Vision encoder ablation.} Language encoder, fusion, and head fixed; encoders fine-tuned except where marked. The VLM rows use only the vision tower of PaliGemma~2 and Qwen3-VL~2B. Cost is per inference with language cached. Bold marks the highest point estimate.}
  \label{tab:abl_vision}
  \begin{tabular*}{\columnwidth}{@{\extracolsep{\fill}} l r r r r @{}}
    \toprule
    Encoder & Params (M) & Succ.\ (\%) & Calls/s & Energy (J) \\
    \midrule
    DINOv3 ViT-B/16 & 86.0 & 54.1 & 145.7 & 2.47 \\
    SigLIP-So400M & 412.4 & 53.9 & 56.7 & 9.04 \\
    Qwen3-VL ViT & 407.0 & 51.8 & 48.8 & 8.31 \\
    DINOv3 ConvNeXt-B, frozen & 87.6 & 32.6 & \textbf{162.7} & \textbf{2.07} \\
    \midrule
    DINOv3 ConvNeXt-B, fine-tuned & 87.6 & \textbf{55.6} & \textbf{162.7} & \textbf{2.07} \\
    \bottomrule
  \end{tabular*}
\end{table}

\subsection{Language encoder ablation}
\label{sec:results_language}

Table~\ref{tab:abl_language} swaps the language encoder with the rest fixed. Two controls bound what language contributes. Without language tokens the policy reaches 11.3\%: in a scene that affords several tasks it performs one of them at random. A one-hot code in place of the instruction raises this to 37.4\%, still 18 points below NeoBERT. The text encoders resolve the task from held-out paraphrases, and the 2019-era encoders trail: T5-base at 44.5\% and BERT-base at
47.8\%, against 50.6\% for mmBERT-base and 55.6\% for NeoBERT. Fine-tuning NeoBERT gives 55.1\%, a difference the single-run design cannot resolve, so DEM keeps it frozen. The two VLM
language models, Gemma~2 from PaliGemma~2 and the Qwen3-VL text model,
reach 55.0\% and 54.3\% with eight to twelve times NeoBERT's parameters and two to three times its encoding latency, so the larger language models do not produce higher observed success on these tasks, and the single-training-run design cannot resolve small differences.

\begin{table}[t]
  \centering
  \scriptsize
  \setlength{\tabcolsep}{3pt}
  \caption{\textbf{Language encoder ablation.} Vision encoder, fusion, and head fixed; encoders frozen except where marked. The VLM rows use only the language model of PaliGemma~2 and Qwen3-VL~2B; None removes the language tokens, One-hot replaces them with a one-hot code. Encode cost is for one standalone encoding, paid once per instruction; Fig.~\ref{fig:architecture} shows the smaller in-pipeline increment. Bold marks the highest point estimate.}
  \label{tab:abl_language}
  \begin{tabular*}{\columnwidth}{@{\extracolsep{\fill}} l r r r r @{}}
    \toprule
    Encoder & Params (M) & Succ.\ (\%) & Encode (ms) & Energy (J) \\
    \midrule
    None & -- & 11.3 & -- & -- \\
    One-hot & -- & 37.4 & -- & -- \\
    T5-base & 109.6 & 44.5 & 2.92 & 0.39 \\
    BERT-base & 109.5 & 47.8 & \textbf{2.18} & \textbf{0.29} \\
    mmBERT-base & 307 & 50.6 & 4.43 & 0.57 \\
    Gemma~2 & 2617 & 55.0 & 16.10 & 5.63 \\
    Qwen3-VL LM & 1721 & 54.3 & 12.06 & 4.01 \\
    NeoBERT, fine-tuned & 222 & 55.1 & 5.15 & 0.97 \\
    \midrule
    NeoBERT, frozen & 222 & \textbf{55.6} & 5.15 & 0.97 \\
    \bottomrule
  \end{tabular*}
\end{table}

\subsection{Action head}
\label{sec:results_action}

Table~\ref{tab:abl_action} swaps the action head with the encoders and fusion fixed. All four heads stay within 3.2 points of one another in observed mean success, whereas their costs differ substantially: ACT and MeanFlow need one pass per chunk and reach about 163 calls per second for about 2.1\,J, while flow matching and Diffusion Policy need ten and sixteen passes and reach 40 and 27 calls per second for 8.5 and 12.9\,J. Because each configuration is trained once, we do not read the small success differences as a ranking. The robust conclusion is computational: MeanFlow reaches the highest point estimate with one head evaluation instead of 10 or 16.
\begin{table}[t]
  \centering
  \scriptsize
  \setlength{\tabcolsep}{3pt}
  \caption{\textbf{Robustness to the action head.} Encoders and fusion fixed; ACT and Diffusion Policy heads matched to MeanFlow in parameter count. Passes is head forward passes per chunk. For reference, the VLM-backbone policies $\pi_{0.5}$ and GR00T N1.7 reach 54.6\% and 56.9\% under the same protocol (Table~\ref{tab:main}). Bold marks the highest decoupled point estimate.}
  \label{tab:abl_action}
  \begin{tabular*}{\columnwidth}{@{\extracolsep{\fill}} l r r r r @{}}
    \toprule
    Action head & Passes & Succ.\ (\%) & Calls/s & Energy (J) \\
    \midrule
    Diffusion Policy & 16 & 53.8 & 26.9 & 12.85 \\
    Flow matching & 10 & 54.7 & 40.3 & 8.50 \\
    ACT & 1 & 52.4 & \textbf{164.4} & 2.10 \\
    \midrule
    MeanFlow & 1 & \textbf{55.6} & 162.7 & \textbf{2.07} \\
    \bottomrule
  \end{tabular*}
\end{table}

\subsection{Robustness beyond MeanFlow}
\label{sec:results_robustness}

Table~\ref{tab:abl_action} tests whether the success of the decoupled
policy depends on MeanFlow in particular. With the DINOv3 and NeoBERT
representation and the cross-attention fusion fixed, flow matching
reaches 54.7\%, Diffusion Policy 53.8\%, and ACT 52.4\%, against 55.6\%
for MeanFlow. For context, the VLM-backbone policies $\pi_{0.5}$ and
GR00T N1.7 reach 54.6\% and 56.9\% under the same protocol
(Table~\ref{tab:main}). The decoupled representation therefore remains
near the VLM-policy range across regression-, diffusion-, and
flow-based action generation, and the system-level comparison with the VLM policies is not explained by the choice of MeanFlow alone.

The VLM baselines also have far greater total capacity and broader
pretraining (Table~\ref{tab:efficiency}), two factors that could
reasonably favor them in downstream transfer, yet they show no clear
success advantage under the trained-task protocol. We read these results
as evidence that modern decoupled components offer a competitive and
substantially more efficient alternative in this regime, not as a
causal claim that joint vision-language encoding is inferior.

\subsection{Chunk length}
\label{sec:results_chunk}

Action chunking exists to amortize slow inference: a policy that needs
100\,ms per call has to commit to many actions per call.
Table~\ref{tab:abl_chunk} sweeps the chunk length from 16 down to a
single action, replanning every $H/2$ steps as Diffusion Policy
does \cite{chi2023diffusionpolicy} and every step for $H = 1$; chunk length and replan interval change together, so the sweep does not separate the two. Success is flat between 16 and 8 (55.6\% and 55.9\%) and declines to 50.6\%
when DEM replans at every control step, while still requiring only one 6.1\,ms policy forward pass per step; that setting still exceeds every policy in Table~\ref{tab:main} except the three at the top. Chunking is therefore not a required component of DEM, although on our tasks the longer chunks remain the better choice.

\begin{table}[t]
  \centering
  \footnotesize
  \caption{\textbf{Chunk length.} $H$ actions per inference, the first $H/2$ executed before replanning \cite{chi2023diffusionpolicy}, every step for $H = 1$. Per-call latency is 6.14\,ms for every $H$.}
  \label{tab:abl_chunk}
  \begin{tabular*}{\columnwidth}{@{\extracolsep{\fill}} l ccccc @{}}
    \toprule
    Chunk length $H$ & 16 & 8 & 4 & 2 & 1 \\
    \midrule
    Replan interval & 8 & 4 & 2 & 1 & 1 \\
    Success (\%) & 55.6 & 55.9 & 52.7 & 51.4 & 50.6 \\
    
    \bottomrule
  \end{tabular*}
\end{table}

\subsection{Real-world tasks}
\label{sec:results_real}

\begin{figure}[t]
  \centering
  \includegraphics[width=0.9\columnwidth]{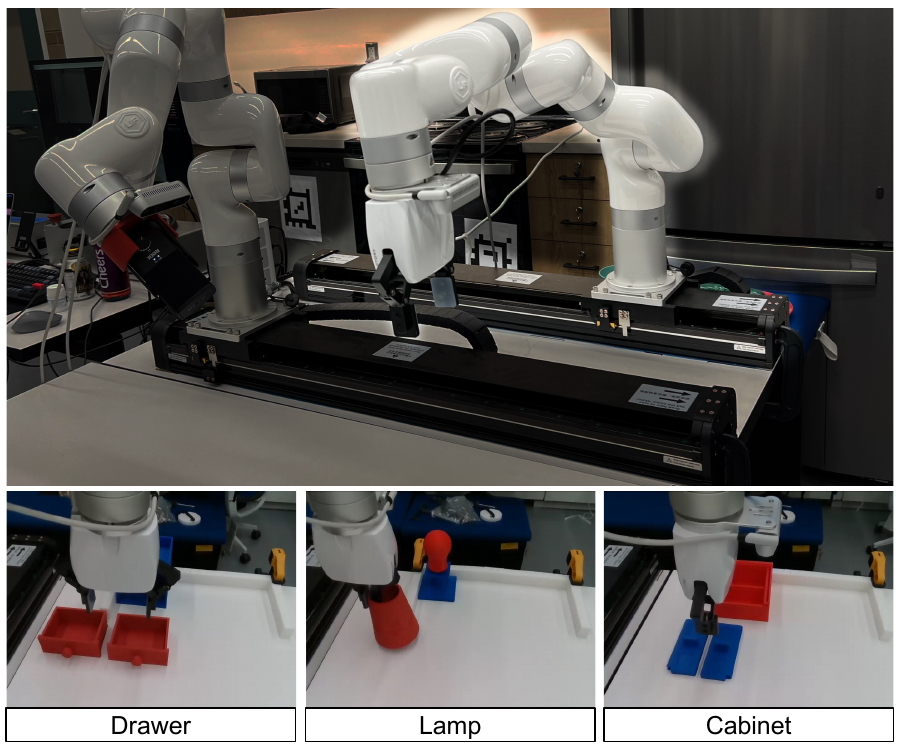}
  \caption{\textbf{Real-world setup and tasks.} Top: an xArm~7 on a linear rail (highlighted), eight degrees of freedom; the second arm only carries the scene camera. Bottom: drawer, lamp, and cabinet assembly adapted from FurnitureBench \cite{heo2023furniturebench}.}
  \label{fig:realworld}
\end{figure}

Table~\ref{tab:real} reports the real-world results on the tasks of
Fig.~\ref{fig:realworld}, with 50 rollouts per task and method. GR00T N1.7 succeeds in 102 of 150 trials (68.0\%) and DEM in 99 (66.0\%), ahead of $\pi_{0.5}$ (63.3\%) and TurboVLA (53.3\%). A three-trial difference cannot support an ordering, particularly with training variability unmeasured, so we take the real-robot study as evidence that DEM's simulated performance transfers to the physical system rather than as evidence of equivalence to GR00T N1.7. The ordering matches the simulation benchmark, where the same three policies finish within 2.3 points of one another. The architectures, inference stack, and workstation are the same as in simulation, so the costs of Table~\ref{tab:efficiency} apply unchanged: DEM reaches this success at 163 policy calls per second and 2.07\,J per call, GR00T N1.7 at 19.3 calls per second and 13.7\,J.

\begin{table}[t]
  \centering
  \footnotesize
    \caption{\textbf{Real-world success rate (\%).} Three tasks of Fig.~\ref{fig:realworld}, 50 rollouts per task and method. Bold marks the highest point estimate.}
  \label{tab:real}
  \begin{tabular*}{\columnwidth}{@{\extracolsep{\fill}} l ccc c @{}}
    \toprule
    Method & Drawer & Lamp & Cabinet & Avg. \\
    \midrule
    $\pi_{0.5}$ & 70 & 80 & 40 & 63.3 \\
    GR00T N1.7 & \textbf{74} & \textbf{84} & \textbf{46} & \textbf{68.0} \\
    TurboVLA & 60 & 72 & 28 & 53.3 \\
    \midrule
    DEM & 72 & 82 & 44 & 66.0 \\
    \bottomrule
  \end{tabular*}
\end{table}

\section{Discussion}
\label{sec:discussion}

On the tasks studied here, DEM provides a better observed success--latency--energy trade-off than the evaluated VLM-backbone policies. Better here means a different point on the observed frontier: GR00T N1.7 has the highest success point estimate, and DEM has far lower measured inference cost, so the results support an efficiency claim rather than a claim that decoupled encoding is more capable than joint vision-language pretraining. Table~\ref{tab:abl_action} shows that its competitive success is not specific to MeanFlow: the same decoupled representation remains near the VLM-policy range with ACT, Diffusion Policy, and flow matching, and MeanFlow turns that representation into inference efficiency by replacing iterative action generation with a single head evaluation. The central empirical finding is that the additional capacity and joint multimodal pretraining of the VLM baselines do not yield a resolved success advantage in the evaluated trained-skill regime. One asymmetry shapes the final design: fine-tuning the vision encoder is worth 23 points, while fine-tuning the language encoder changes nothing, since the workspace, the
gripper, and the objects are specific to the robot and absent from web
pretraining, whereas the instructions are short and their vocabulary is ordinary English.

\textbf{Limitations.} (i) Trained-task scope: every
evaluated skill appears in the demonstrations, and no rollout asks for
an object, a task, or a compositional instruction absent from training,
so the experiments do not test whether joint vision-language
pretraining helps beyond the trained skills. (ii) System-level
comparison: the VLM baselines keep their native fusion, action modules,
pretraining, and optimization recipes, so the results describe the
observed success--cost trade-off and do not isolate the effect of joint
versus decoupled encoding. (iii) Training variability: each
configuration is trained once, so the reported standard errors cover
evaluation sampling only. (iv) Device-level efficiency: latency and
GPU energy measure the policy's forward pass on one workstation GPU,
not end-to-end control frequency or system energy. These boundaries mark where the evidence is strongest:
repeated execution of a known collection of language-conditioned
skills, for which the forward-pass cost is paid throughout deployment.

\section{Conclusion}
\label{sec:conclusion}

Under the evaluated trained-task protocol, DEM reaches an observed
success range similar to the strongest VLM-backbone policies while
requiring far less forward-pass computation. The results do not show
that decoupled encoding is preferable in general, and they do not test
the open-world generalization for which joint vision-language
pretraining may matter most. They establish an efficiency baseline: for
repeatedly executed skills represented in the demonstrations, a modern
decoupled policy can occupy a favorable point on the
success--latency--energy frontier. Future comparisons on unseen skills, objects, and compositional instructions should test whether the cost of a jointly fused VLM backbone buys benefits that the trained-task regime does not reveal.

\section*{ACKNOWLEDGMENT}
The instruction paraphrases used for training and evaluation
(Section~\ref{sec:setup}) were generated with Claude
Opus~4.8 \cite{anthropic2026claude}.

\bibliographystyle{IEEEtranN}
\bibliography{refs}

@IEEEtranBSTCTL{IEEEexample:BSTcontrol,
  CTLuse_forced_etal       = {yes},
  CTLmax_names_forced_etal = {6},
  CTLnames_show_etal       = {1}
}

@inproceedings{brohan2022rt1,
  title     = {{RT-1}: Robotics Transformer for Real-World Control at Scale},
  author    = {Brohan, Anthony and Brown, Noah and Carbajal, Justice and Chebotar, Yevgen and Dabis, Joseph and Finn, Chelsea and Gopalakrishnan, Keerthana and Hausman, Karol and Herzog, Alex and Hsu, Jasmine and Ibarz, Julian and Ichter, Brian and Irpan, Alex and Jackson, Tomas and Jesmonth, Sally and Joshi, Nikhil J. and Julian, Ryan and Kalashnikov, Dmitry and Kuang, Yuheng and Leal, Isabel and Lee, Kuang-Huei and Levine, Sergey and Lu, Yao and Malla, Utsav and Manjunath, Deeksha and Mordatch, Igor and Nachum, Ofir and Parada, Carolina and Peralta, Jodilyn and Perez, Emily and Pertsch, Karl and Quiambao, Jornell and Rao, Kanishka and Ryoo, Michael and Salazar, Grecia and Sanketi, Pannag and Sayed, Kevin and Singh, Jaspiar and Sontakke, Sumedh and Stone, Austin and Tan, Clayton and Tran, Huong and Vanhoucke, Vincent and Vega, Steve and Vuong, Quan and Xia, Fei and Xiao, Ted and Xu, Peng and Xu, Sichun and Yu, Tianhe and Zitkovich, Brianna},
  booktitle = {Robotics: Science and Systems (RSS)},
  year      = {2023}
}

@inproceedings{octo2024octo,
  title     = {Octo: An Open-Source Generalist Robot Policy},
  author    = {{Octo Model Team} and Ghosh, Dibya and Walke, Homer and Pertsch, Karl and Black, Kevin and Mees, Oier and Dasari, Sudeep and Hejna, Joey and Kreiman, Tobias and Xu, Charles and Luo, Jianlan and Tan, You Liang and Chen, Lawrence Yunliang and Sanketi, Pannag and Vuong, Quan and Xiao, Ted and Sadigh, Dorsa and Finn, Chelsea and Levine, Sergey},
  booktitle = {Robotics: Science and Systems (RSS)},
  year      = {2024}
}

@inproceedings{kim2024openvla,
  title     = {{OpenVLA}: An Open-Source Vision-Language-Action Model},
  author    = {Kim, Moo Jin and Pertsch, Karl and Karamcheti, Siddharth and Xiao, Ted and Balakrishna, Ashwin and Nair, Suraj and Rafailov, Rafael and Foster, Ethan and Lam, Grace and Sanketi, Pannag and Vuong, Quan and Kollar, Thomas and Burchfiel, Benjamin and Tedrake, Russ and Sadigh, Dorsa and Levine, Sergey and Liang, Percy and Finn, Chelsea},
  booktitle = {Conference on Robot Learning (CoRL)},
  year      = {2024}
}

@inproceedings{black2024pi0,
  title     = {{$\pi_0$}: A Vision-Language-Action Flow Model for General Robot Control},
  author    = {Black, Kevin and Brown, Noah and Driess, Danny and Esmail, Adnan and Equi, Michael and Finn, Chelsea and Fusai, Niccolo and Groom, Lachy and Hausman, Karol and Ichter, Brian and Jakubczak, Szymon and Jones, Tim and Ke, Liyiming and Levine, Sergey and Li-Bell, Adrian and Mothukuri, Mohith and Nair, Suraj and Pertsch, Karl and Shi, Lucy Xiaoyang and Tanner, James and Vuong, Quan and Walling, Anna and Wang, Haohuan and Zhilinsky, Ury},
  booktitle = {Robotics: Science and Systems (RSS)},
  year      = {2025}
}

@article{black2025pi05,
  title   = {{$\pi_{0.5}$}: a Vision-Language-Action Model with Open-World Generalization},
  author  = {{Physical Intelligence} and Black, Kevin and Brown, Noah and Darpinian, James and Dhabalia, Karan and Driess, Danny and Esmail, Adnan and Equi, Michael and Finn, Chelsea and Fusai, Niccolo and Galliker, Manuel Y. and Ghosh, Dibya and Groom, Lachy and Hausman, Karol and Ichter, Brian and Jakubczak, Szymon and Jones, Tim and Ke, Liyiming and LeBlanc, Devin and Levine, Sergey and Li-Bell, Adrian and Mothukuri, Mohith and Nair, Suraj and Pertsch, Karl and Ren, Allen Z. and Shi, Lucy Xiaoyang and Smith, Laura and Springenberg, Jost Tobias and Stachowicz, Kyle and Tanner, James and Vuong, Quan and Walke, Homer and Walling, Anna and Wang, Haohuan and Yu, Lili and Zhilinsky, Ury},
  journal = {arXiv preprint arXiv:2504.16054},
  year    = {2025}
}

@article{bjorck2025gr00tn1,
  title   = {{GR00T N1}: An Open Foundation Model for Generalist Humanoid Robots},
  author  = {Bjorck, Johan and Casta{\~n}eda, Fernando and Cherniadev, Nikita and Da, Xingye and Ding, Runyu and Fan, Linxi and Fang, Yu and Fox, Dieter and Hu, Fengyuan and Huang, Spencer and Jang, Joel and Jiang, Zhenyu and Kautz, Jan and Kundalia, Kaushil and Lao, Lawrence and Li, Zhiqi and Lin, Zongyu and Lin, Kevin and Liu, Guilin and Llontop, Edith and Magne, Loic and Mandlekar, Ajay and Narayan, Avnish and Nasiriany, Soroush and Reed, Scott and Tan, You Liang and Wang, Guanzhi and Wang, Zu and Wang, Jing and Wang, Qi and Xiang, Jiannan and Xie, Yuqi and Xu, Yinzhen and Xu, Zhenjia and Ye, Seonghyeon and Yu, Zhiding and Zhang, Ao and Zhang, Hao and Zhao, Yizhou and Zheng, Ruijie and Zhu, Yuke},
  journal = {arXiv preprint arXiv:2503.14734},
  year    = {2025}
}

@misc{nvidia2026gr00tn17,
  title        = {{GR00T-N1.7-3B} Model Card},
  author       = {{NVIDIA}},
  year         = {2026},
  howpublished = {\url{https://huggingface.co/nvidia/GR00T-N1.7-3B}},
  note         = {{Hugging Face} model card, first published 2026-02-25, accessed 2026-09-14}
}

@inproceedings{zhao2023act,
  title     = {Learning Fine-Grained Bimanual Manipulation with Low-Cost Hardware},
  author    = {Zhao, Tony Z. and Kumar, Vikash and Levine, Sergey and Finn, Chelsea},
  booktitle = {Robotics: Science and Systems (RSS)},
  year      = {2023}
}

@inproceedings{kim2025oft,
  title     = {Fine-Tuning Vision-Language-Action Models: Optimizing Speed and Success},
  author    = {Kim, Moo Jin and Finn, Chelsea and Liang, Percy},
  booktitle = {Robotics: Science and Systems (RSS)},
  year      = {2025}
}

@inproceedings{black2025rtc,
  title     = {Real-Time Execution of Action Chunking Flow Policies},
  author    = {Black, Kevin and Galliker, Manuel Y. and Levine, Sergey},
  booktitle = {Advances in Neural Information Processing Systems (NeurIPS)},
  year      = {2025}
}

@article{wen2024tinyvla,
  title   = {{TinyVLA}: Toward Fast, Data-Efficient Vision-Language-Action Models for Robotic Manipulation},
  author  = {Wen, Junjie and Zhu, Yichen and Li, Jinming and Zhu, Minjie and Tang, Zhibin and Wu, Kun and Xu, Zhiyuan and Liu, Ning and Cheng, Ran and Shen, Chaomin and Peng, Yaxin and Feng, Feifei and Tang, Jian},
  journal = {IEEE Robotics and Automation Letters},
  volume  = {10},
  number  = {4},
  pages   = {3988--3995},
  year    = {2025},
  doi     = {10.1109/LRA.2025.3544909}
}

@article{shukor2025smolvla,
  title   = {{SmolVLA}: A Vision-Language-Action Model for Affordable and Efficient Robotics},
  author  = {Shukor, Mustafa and Aubakirova, Dana and Capuano, Francesco and Kooijmans, Pepijn and Palma, Steven and Zouitine, Adil and Aractingi, Michel and Pascal, Caroline and Russi, Martino and Marafioti, Andres and Alibert, Simon and Cord, Matthieu and Wolf, Thomas and Cadene, Remi},
  journal = {arXiv preprint arXiv:2506.01844},
  year    = {2025}
}

@article{xu2025vlacache,
  title   = {{VLA-Cache}: Efficient Vision-Language-Action Manipulation via Adaptive Token Caching},
  author  = {Xu, Siyu and Wang, Yunke and Xia, Chenghao and Zhu, Dihao and Huang, Tao and Xu, Chang},
  journal = {arXiv preprint arXiv:2502.02175},
  year    = {2025}
}

@inproceedings{yue2024deervla,
  title     = {{DeeR-VLA}: Dynamic Inference of Multimodal Large Language Models for Efficient Robot Execution},
  author    = {Yue, Yang and Wang, Yulin and Kang, Bingyi and Han, Yizeng and Wang, Shenzhi and Song, Shiji and Feng, Jiashi and Huang, Gao},
  booktitle = {Advances in Neural Information Processing Systems (NeurIPS)},
  year      = {2024}
}

@article{xie2026turbovla,
  title   = {{TurboVLA}: Real-Time Vision-Language-Action Model at 32 {Hz} on an {RTX} 4090 with $<$1 {GB} {VRAM}},
  author  = {Xie, Hengyi and Yao, Chenfei and Wu, Xianjin and Zhu, Yingying and Liang, Dingkang and Bai, Xiang and Ding, Han},
  journal = {arXiv preprint arXiv:2607.27205},
  year    = {2026}
}

@inproceedings{chi2023diffusionpolicy,
  title     = {Diffusion Policy: Visuomotor Policy Learning via Action Diffusion},
  author    = {Chi, Cheng and Xu, Zhenjia and Feng, Siyuan and Cousineau, Eric and Du, Yilun and Burchfiel, Benjamin and Tedrake, Russ and Song, Shuran},
  booktitle = {Robotics: Science and Systems (RSS)},
  year      = {2023}
}

@inproceedings{geng2025meanflow,
  title     = {Mean Flows for One-step Generative Modeling},
  author    = {Geng, Zhengyang and Deng, Mingyang and Bai, Xingjian and Kolter, J. Zico and He, Kaiming},
  booktitle = {Advances in Neural Information Processing Systems (NeurIPS)},
  year      = {2025}
}

@inproceedings{sheng2025mp1,
  title     = {{MP1}: {MeanFlow} Tames Policy Learning in 1-step for Robotic Manipulation},
  author    = {Sheng, Juyi and Wang, Ziyi and Li, Peiming and Liu, Mengyuan},
  booktitle = {AAAI Conference on Artificial Intelligence (AAAI)},
  year      = {2026}
}

@inproceedings{nasiriany2024robocasa,
  title     = {{RoboCasa}: Large-Scale Simulation of Everyday Tasks for Generalist Robots},
  author    = {Nasiriany, Soroush and Maddukuri, Abhiram and Zhang, Lance and Parikh, Adeet and Lo, Aaron and Joshi, Abhishek and Mandlekar, Ajay and Zhu, Yuke},
  booktitle = {Robotics: Science and Systems (RSS)},
  year      = {2024}
}

@inproceedings{devlin2019bert,
  title     = {{BERT}: Pre-training of Deep Bidirectional Transformers for Language Understanding},
  author    = {Devlin, Jacob and Chang, Ming-Wei and Lee, Kenton and Toutanova, Kristina},
  booktitle = {Conference of the North American Chapter of the Association for Computational Linguistics (NAACL-HLT)},
  year      = {2019}
}

@article{warner2024modernbert,
  title   = {Smarter, Better, Faster, Longer: A Modern Bidirectional Encoder for Fast, Memory Efficient, and Long Context Finetuning and Inference},
  author  = {Warner, Benjamin and Chaffin, Antoine and Clavi{\'e}, Benjamin and Weller, Orion and Hallstr{\"o}m, Oskar and Taghadouini, Said and Gallagher, Alexis and Biswas, Raja and Ladhak, Faisal and Aarsen, Tom and Cooper, Nathan and Adams, Griffin and Howard, Jeremy and Poli, Iacopo},
  journal = {arXiv preprint arXiv:2412.13663},
  year    = {2024}
}

@article{marone2025mmbert,
  title   = {{mmBERT}: A Modern Multilingual Encoder with Annealed Language Learning},
  author  = {Marone, Marc and Weller, Orion and Fleshman, William and Yang, Eugene and Lawrie, Dawn and Van Durme, Benjamin},
  journal = {arXiv preprint arXiv:2509.06888},
  year    = {2025}
}

@article{lebreton2025neobert,
  title   = {{NeoBERT}: A Next-Generation {BERT}},
  author  = {Le Breton, Lola and Fournier, Quentin and El Mezouar, Mariam and Morris, John X. and Chandar, Sarath},
  journal = {arXiv preprint arXiv:2502.19587},
  year    = {2025}
}

@inproceedings{weller2026ettin,
  title     = {Seq vs Seq: An Open Suite of Paired Encoders and Decoders},
  author    = {Weller, Orion and Ricci, Kathryn and Marone, Marc and Chaffin, Antoine and Lawrie, Dawn and Van Durme, Benjamin},
  booktitle = {International Conference on Learning Representations (ICLR)},
  year      = {2026}
}

@article{simeoni2025dinov3,
  title   = {{DINOv3}},
  author  = {Sim{\'e}oni, Oriane and Vo, Huy V. and Seitzer, Maximilian and Baldassarre, Federico and Oquab, Maxime and Jose, Cijo and Khalidov, Vasil and Szafraniec, Marc and Yi, Seungeun and Ramamonjisoa, Micha{\"e}l and Massa, Francisco and Haziza, Daniel and Wehrstedt, Luca and Wang, Jianyuan and Darcet, Timoth{\'e}e and Moutakanni, Th{\'e}o and Sentana, Leonel and Roberts, Claire and Vedaldi, Andrea and Tolan, Jamie and Brandt, John and Couprie, Camille and Mairal, Julien and J{\'e}gou, Herv{\'e} and Labatut, Patrick and Bojanowski, Piotr},
  journal = {arXiv preprint arXiv:2508.10104},
  year    = {2025}
}

@inproceedings{brown2020gpt3,
  title     = {Language Models are Few-Shot Learners},
  author    = {Brown, Tom B. and Mann, Benjamin and Ryder, Nick and Subbiah, Melanie and Kaplan, Jared and Dhariwal, Prafulla and Neelakantan, Arvind and Shyam, Pranav and Sastry, Girish and Askell, Amanda and Agarwal, Sandhini and Herbert-Voss, Ariel and Krueger, Gretchen and Henighan, Tom and Child, Rewon and Ramesh, Aditya and Ziegler, Daniel M. and Wu, Jeffrey and Winter, Clemens and Hesse, Christopher and Chen, Mark and Sigler, Eric and Litwin, Mateusz and Gray, Scott and Chess, Benjamin and Clark, Jack and Berner, Christopher and McCandlish, Sam and Radford, Alec and Sutskever, Ilya and Amodei, Dario},
  booktitle = {Advances in Neural Information Processing Systems (NeurIPS)},
  year      = {2020}
}

@inproceedings{majumdar2023vc1,
  title     = {Where are we in the search for an Artificial Visual Cortex for Embodied Intelligence?},
  author    = {Majumdar, Arjun and Yadav, Karmesh and Arnaud, Sergio and Ma, Yecheng Jason and Chen, Claire and Silwal, Sneha and Jain, Aryan and Berges, Vincent-Pierre and Abbeel, Pieter and Malik, Jitendra and Batra, Dhruv and Lin, Yixin and Maksymets, Oleksandr and Rajeswaran, Aravind and Meier, Franziska},
  booktitle = {Advances in Neural Information Processing Systems (NeurIPS)},
  year      = {2023}
}

@article{sun2025prismdp,
  title   = {{PRISM-DP}: Spatial Pose-based Observations for Diffusion-Policies via Segmentation, Mesh Generation, and Pose Tracking},
  author  = {Sun, Xiatao and Chen, Yinxing and Rakita, Daniel},
  journal = {arXiv preprint arXiv:2504.20359},
  year    = {2025}
}

@inproceedings{sun2025dynamic,
  title     = {Dynamic Rank Adjustment in Diffusion Policies for Efficient and Flexible Training},
  author    = {Sun, Xiatao and Yang, Shuo and Chen, Yinxing and Fan, Francis and Liang, Yiyan and Rakita, Daniel},
  booktitle = {Robotics: Science and Systems (RSS)},
  year      = {2025}
}

@inproceedings{sun2026hybrid,
  title     = {Hybrid Diffusion Policies with Projective Geometric Algebra for Efficient Robot Manipulation Learning},
  author    = {Sun, Xiatao and Wang, Yuxuan and Yang, Shuo and Chen, Yinxing and Rakita, Daniel},
  booktitle = {IEEE International Conference on Robotics and Automation (ICRA)},
  year      = {2026}
}

@article{sun2024optimizing,
  title   = {Optimizing Active Perception for Learning Simultaneous Viewpoint Selection and Manipulation with Diffusion Policy},
  author  = {Sun, Xiatao and Fan, Francis and Chen, Yinxing and Rakita, Daniel},
  journal = {arXiv preprint arXiv:2409.14615},
  year    = {2024}
}

@article{sun2026artificial,
  title   = {Artificial Foveated Perception for Mitigating Shortcut Learning in Robotic Foundation Models},
  author  = {Sun, Xiatao and Zhuang, Yuan and Sanchez Lopez Negrete, Mateo and Coldea, Matei-Victor and Liang, Chen and Zhang, Haoyang and Liu, Che and Zeng, Ziyao and Li, Shawn and Wang, Qian and Miao, Fei and Rakita, Daniel},
  journal = {arXiv preprint arXiv:2607.10655},
  year    = {2026}
}

@inproceedings{wang2025subsecond,
  title     = {Subsecond {3D} Mesh Generation for Robot Manipulation},
  author    = {Wang, Qian and Abdellall, Omar and Gao, Tony and Sun, Xiatao and Rakita, Daniel},
  booktitle = {IEEE International Conference on Robotics and Automation (ICRA)},
  year      = {2026}
}

@inproceedings{tong2024cambrian,
  title     = {{Cambrian-1}: A Fully Open, Vision-Centric Exploration of Multimodal {LLMs}},
  author    = {Tong, Shengbang and Brown, Ellis and Wu, Penghao and Woo, Sanghyun and Middepogu, Manoj and Akula, Sai Charitha and Yang, Jihan and Yang, Shusheng and Iyer, Adithya and Pan, Xichen and Wang, Ziteng and Fergus, Rob and LeCun, Yann and Xie, Saining},
  booktitle = {Advances in Neural Information Processing Systems (NeurIPS)},
  year      = {2024}
}

@article{sendai2026minerva,
  title   = {{MINERVA}: How Small Can a Manipulation Policy Be and Still Solve {LIBERO}?},
  author  = {Sendai, Kohei and Matsushima, Tatsuya and Iwasawa, Yusuke},
  journal = {arXiv preprint arXiv:2609.03715},
  year    = {2026}
}

@inproceedings{peebles2023dit,
  title     = {Scalable Diffusion Models with Transformers},
  author    = {Peebles, William and Xie, Saining},
  booktitle = {IEEE/CVF International Conference on Computer Vision (ICCV)},
  year      = {2023}
}

@article{chen2026meanflowvla,
  title   = {Mean-Flow based One-Step Vision-Language-Action},
  author  = {Chen, Yang and Ma, Xiaoguang and Zhao, Bin},
  journal = {arXiv preprint arXiv:2603.01469},
  year    = {2026}
}

@article{guo2026reactvla,
  title   = {{ReactVLA}: Fast and Lightweight Reactive Robot Manipulation via Improved Mean Flow Action Generation},
  author  = {Guo, Yanzhao and Chen, Wenkai and Zhang, Jianwei},
  journal = {arXiv preprint arXiv:2606.14255},
  year    = {2026}
}

@misc{nvidia2025kitchendemos,
  title        = {{PhysicalAI-Robotics-Manipulation-Kitchen-Demos}},
  author       = {{NVIDIA}},
  year         = {2025},
  howpublished = {\url{https://huggingface.co/datasets/nvidia/PhysicalAI-Robotics-Manipulation-Kitchen-Demos}},
  note         = {{Hugging Face} dataset, CC-BY-4.0, accessed 2026-09-14}
}

@inproceedings{heo2023furniturebench,
  title     = {{FurnitureBench}: Reproducible Real-World Benchmark for Long-Horizon Complex Manipulation},
  author    = {Heo, Minho and Lee, Youngwoon and Lee, Doohyun and Lim, Joseph J.},
  booktitle = {Robotics: Science and Systems (RSS)},
  year      = {2023}
}

@article{haldar2024baku,
  title   = {{BAKU}: An Efficient Transformer for Multi-Task Policy Learning},
  author  = {Haldar, Siddhant and Peng, Zhuoran and Pinto, Lerrel},
  journal = {arXiv preprint arXiv:2406.07539},
  year    = {2024}
}

@misc{anthropic2026claude,
  author       = {{Anthropic}},
  title        = {{Claude Opus 4.8}},
  howpublished = {\url{https://www.anthropic.com/claude}},
  year         = {2026},
  note         = {{Large language model}, accessed 2026-09-15}
}

@article{chen2026letitbesimple,
  title   = {Let It Be Simple: One-Step Action Generation for Vision-Language-Action Models},
  author  = {Chen, Yitong and Zhang, Shiduo and Gong, Jingjing and Qiu, Xipeng},
  journal = {arXiv preprint arXiv:2606.05737},
  year    = {2026}
}

@article{vanjani2026damvla,
  title   = {{DAM-VLA}: Decoupled Asynchronous Multimodal Vision Language Action Model},
  author  = {Vanjani, Pankhuri and Li, Zhuoyue and Suliga, Jakub and Reuss, Moritz and Geraci, Gianluca and Jiang, Xinkai and Lioutikov, Rudolf},
  journal = {arXiv preprint arXiv:2606.12105},
  year    = {2026}
}

\end{document}